\documentclass[fullpage]{article}

\usepackage[margin=2.2cm]{geometry}
\usepackage{graphicx}
\usepackage{url}
\usepackage{enumitem}
\usepackage{multicol}
\usepackage{multirow}
\usepackage{float}
\usepackage{eurosym}

\newcommand{\W}{$\bullet$}
\newcommand{\D}{--}

\begin{document}
\title{Second International Nurse Rostering Competition (INRC-II)\\
--- Problem Description and Rules ---}
\author{Sara Ceschia$^1$,
Nguyen Thi Thanh Dang$^2$,
Patrick De Causmaecker$^2$,\\
Stefaan Haspeslagh$^3$,
Andrea Schaerf$^1$\\[5mm]
\normalsize 1. DIEGM, University of Udine\\
\normalsize via delle Scienze 206, 33100, Udine, Italy\\
\normalsize \url{{sara.ceschia,schaerf}@uniud.it}\\[2mm]
\normalsize 2. KU Leuven, Department of Computer Science, CODeS \& iMinds-ITEC, KULAK\\
\normalsize E. Sabbelaan 53, 8500 Kortrijk, Belgium\\
\normalsize \url{{nguyenthithanh.dang, patrick.decausmaecker}@kuleuven-kulak.be}\\[2mm]
\normalsize 3. Vives University College, Commercial sciences and business management, MoBiz\\ 
\normalsize Doorniksesteenweg 145, 8500 Kortrijk, Belgium\\ 
\normalsize \url{stefaan.haspeslagh@vives.be}
}

\maketitle

\begin{abstract}
  In this paper, we provide all information to participate to the Second
  International Nurse Rostering Competition (INRC-II). First, we
  describe the problem formulation, which, differently from INRC-I, is
  a multi-stage procedure. Second, we illustrate all the necessary
  infrastructure do be used together with the participant's solver,
  including the testbed, the file formats, and the
  validation/simulation tools. Finally, we state the rules of the
  competition. All update-to-date information about the competition is
  available at \url{http://mobiz.vives.be/inrc2/}.
\end{abstract}

\section{Introduction}

Nurse rostering is a very important problem in healthcare management.
Early papers date from the seventies, but especially in the last
decade, it has drawn significant attention; see \cite{BDVV04,DV11} for
a review of literature and a classification.

The First International Nurse Rostering Competition (INRC-I)
\cite{HDSS12} was run in 2010. The competition welcomed 15 submissions
in three categories (sprint, medium and long tracks).  Since then,
several research groups took this formulation and the corresponding
instances as a challenge
\cite{Awadallah2011,Burke2014,DCS12,Geiger2011,Santos2012,LUAho12,Solos2013}
and produced remarkable results. Optimal solutions as well as new best
solutions have also been found and reported.

The problem considered for INRC-I was the assignment of nurses to
shifts in a fixed planning horizon, subject to a large number of hard
and soft constraint types.

For the Second International Nurse Rostering Competition (INRC-II), we
propose a smaller set of constraint types, but within a
\emph{multi-stage} formulation of the problem. That is, the solvers of
the participants are requested to deal with a sequence of cases,
referring to consecutive weeks of a longer planning horizon (4 or 8
weeks). 

The search method designed by the participants has to be able to solve
a single stage of the problem corresponding to one week. Some
information, called \emph{history}, is carried out between consecutive
weeks, so that the one coming from the previous week has to be taken
into account by the solver. The history includes \emph{border} data,
such as the last worked shift of each nurse, and \emph{counters} for
cumulative data, such as total worked night shifts. Counters' value
has to be checked against global thresholds, but only at the end of
the planning period. The planning horizon is not \emph{rolling}
\cite{BaPu2005} but fixed, in the sense that in the final week all
counters are checked against their limits.

We provide a simple command-line simulation/validation software to be
used simulate the solution process and to evaluate the quality of the
solver. The \emph{simulator} invokes the participant's solver for each
stage iteratively, then updates the history after each single
execution.  The provided \emph{validator} concatenates the solutions
for all weeks, and evaluates them all together, along with the
cumulative data coming from the final history.

The solver should take into account the following separate input
sources:
\begin{description}
\item[Scenario:] Information that is global to all weeks of
  the entire planning horizon, such as nurse contracts and shift types.
\item[Week data:] Specific data of the single week,
  like daily coverage requirements and nurse preferences for specific
  days.
\item[History:] Information that must be passed from a week
  to the other, in order to compute constraint violations properly. It
  includes border information and global counters.
\end{description}

The solver must deliver an output file, based on which, the simulator
computes the new history file, to be passed back to the solver for the
solution of the next week. As will be explained further, besides the 
mandatory input and output files, a custom data file may be used to 
exchange information between two stages.

The paper is organised as follows. Section~\ref{sec:definition}
illustrates the problem definition. Section~\ref{sec:instances}
describes the testbed.  Section~\ref{sec:tools} shows the software
tools made available to the participants to evaluate their solver.
Finally, Section~\ref{sec:rules} describes the rules of the
competition. Appendix~\ref{sec:formats} describes the file formats and
Appendix~\ref{sec:evaluation} provides a deeper look at the constraint
evaluation.  All update-to-date information about the competition is
available at \url{http://mobiz.vives.be/inrc2/}.

\section{Problem definition}
\label{sec:definition}

The basic (one-stage) problem consists in the weekly scheduling of a
fixed number of nurses using a set of shifts, such that in each day a
nurse works a shift or has a day-off. Nurses may have multiple skills,
and for each skill we are given different coverage requirements.

Given the multi-stage nature of the overall process, the input data of
the problem comes from three different sources, called scenario, week
data, and history, as explained in the following sections. The way the
information is organised in the files is explained in
Appendix~\ref{sec:formats}.

\subsection{Scenario}

The scenario represents the general data common to all stages of the
overall process.  It contains the following information:

\begin{description}
\item[Planning horizon:] The number of weeks that compose the planning
  period.
\item[Skills:] The list of skills included in the problem (head nurse,
  regular nurse, trainee, \dots). Each nurse has one or more skills,
  but in each working shift she/he covers exactly one skill request.
\item[Contracts:] Each nurse has one specific contract (full
  time, part time, on call, \dots). The contract sets limits on the
  distribution and the number of assignments within the planning
  horizon. In detail, it contains:
\begin{itemize}
\item minimum and maximum total number of assignments in the planning horizon;
\item minimum and maximum number of consecutive working days;
\item minimum and maximum number of consecutive days-off;
\item maximum number of working week-ends in the planning horizon;
\item a Boolean value representing the presence of the \emph{Complete
    week-end} constraint to the nurse, which states that the nurse
  should work both days of the week-end or none of them.
\end{itemize}
\item[Nurses:] For each nurse, the name (identifier), the
  contract and the set of skills are given.
\item[Shift types:] For each shift type (early, late, night, \dots),
  it is given the minimum and maximum number of consecutive
  assignments of that specific type, and a matrix of forbidden shift type
  successions is given. For example, it may not be allowed to assign
  to a nurse an early shift the day after a late one.
\end{description}

\subsection{Week data}

The week data contains the specific data for the single week. It
consists of the following information:
    
\begin{description}
\item[Requirements:] It is given, for each shift, for each skill, for each week
  day, the optimal and minimum number of nurses necessary to fulfil
  the working duties.
\item[Nurse requests:] It is given, a set of triples, each one
  composed by the nurse name, the week day, and a shift. The presence
  of a given triple represents the request of the nurse not to work in
  the given shift in the given day. The special shift name
  \texttt{Any} represents the request of not working in any shift of
  the day, i.e.  having a day-off.
\end{description}

The above information varies from week to week, due to variability on
the number of current patients and specific preferences of nurses.

Conventionally, all weeks start with Monday, so that the data is
stored in the order \texttt{Mon}, \texttt{Tue}, \dots, \texttt{Sun}.
 
\subsection{History}

The history contains the information that must be carried over from one
week to the following one, so as to evaluate the constraints
correctly. In detail, it reports for each nurse the following two types of information:

\begin{description}
\item[Border data:] The border data is used for checking the
  constraints on consecutive assignments. They are:
  \begin{itemize}
  \item shift worked in the last day of the previous week, or
    the special value \texttt{None} if the nurse had a day-off;
  \item number of consecutive worked shifts of the same type and number of
    consecutive worked shifts in general (both are 0 if the last worked shift
    is \texttt{None});
  \item number of consecutive days-off (0 if the last worked shift is
    not \texttt{None}).
  \end{itemize} 
\item[Counters:] The counters collect the cumulative value over the
  weeks of specific quantities of interest, that have to be checked only at
  the end of the planning period. They are:
  \begin{itemize}
  \item total number of worked shifts;
  \item total number of worked week-ends.
  \end{itemize} 
\end{description}

The history is computed after each week based on the solution delivered
by the solver and on the previous history.  The history passed to the
solver for the first week has all counter values equal to 0, whereas
the border data can have any value.

\subsection{Solution}
\label{subsec:full_solution_process}

The full solution process is a loop that executes at each step the
solver for a week, iterating for all weeks in the planning period (4
or 8 weeks).

After each week, the history information is computed based on the
solution and the previous history, and delivered in a new file. This
is done by the simulator provided, whereas the
solver must deliver only the solution itself.

\begin{figure}[htbp]
\caption{The overall solution/simulation/validation process (4 weeks).}
\label{fig:overall-process}
\begin{center}
\includegraphics[width=\textwidth]{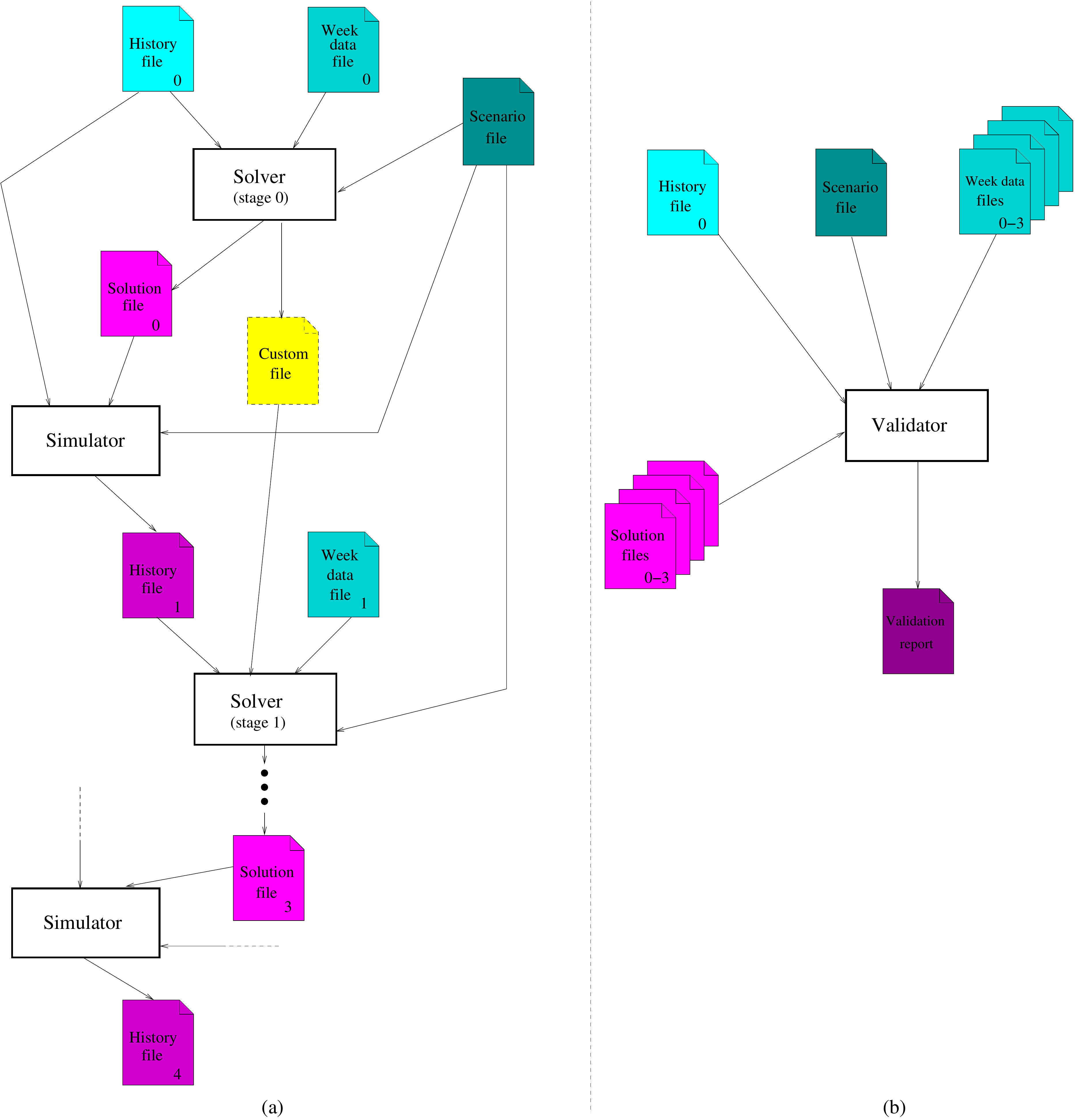}
\end{center}
\end{figure}

The complete process is sketched in Figure~\ref{fig:overall-process}(a),
assuming a scenario of 4 weeks. Input files are colored in different
shades of cyan, and the output ones in different shades on magenta.

In addition to generating a solution file after solving each stage,
the solver might want to save some other prediction information in a
custom file (in yellow in Figure~\ref{fig:overall-process}) and
passing it to the next solver call, in order to guide the solving of
the next stage better. The content and the format of this file is
free, but its name is set by the simulator, as described in
Section~\ref{sec:simulator}.

The output produced by the solver is a list of assignments of nurses to
shifts and skills. Each entry contains the nurse name, the week day,
the shift, and the skill. As an example, consider the entry
$\langle$\texttt{Mary}, \texttt{Tue}, \texttt{Night},
\texttt{HeadNurse}$\rangle$, that states that the nurse \texttt{Mary}
works on Tuesday the night shift with the role of head nurse.

The quality of the overall solution is evaluated, as shown in
Figure~\ref{fig:overall-process}(b), for the entire planning horizon
based on:
\begin{itemize}
\item the solution for each week containing the assignments of the
  nurses to shifts and skills, using the requirements in the week data
  file and the border data in the history file;
\item the counters of the \emph{final} history file, against the limits
  provided in the scenario.
\end{itemize}

\subsection{Constraints}
\label{sec:constraints}

According to the setting outlined above, we split the constraints into two sets: those that can be
computed for each week separately, and those that are computed only
globally at the end of the planning period.

As customary, they are also split into hard and soft constraints.
The former must be always satisfied, and the latter contribute to
the objective function. The weight of each single soft constraint is shown prior to its description below.

\subsubsection{Constraints on the single week}
\label{sec:week-constraints}

Below is the list of hard (H) and soft (S) constraint types:
\begin{description}
\item[\textsf{H1}. Single assignment per day:] A nurse can be assigned to at most one shift per day.
\item[\textsf{H2}. Under-staffing:] The number of nurses for each shift for each
  skill must be at least equal to the minimum requirement.  
\item[\textsf{H3}. Shift type successions:] The shift type assignments of one nurse in two consecutive days
must belong to the legal successions provided in the scenario.  
\item[\textsf{H4}. Missing required skill:] A shift of a given skill must necessarily be fulfilled
  by a nurse having that skill.
\item[\textsf{S1}. Insufficient staffing for optimal coverage (30):] The number of nurses for each shift for each
  skill must be equal to the optimal requirement.  Each missing nurse is penalised 
  according to the weight provided. Extra nurses
  above the optimal value are not considered in the cost.
\item[\textsf{S2}. Consecutive assignments (15/30):] Minimum and maximum number
  of consecutive assignments, per shift or global, should
  be respected.  Their evaluation involves also the border data. Each
  extra or missing day is multiplied by the corresponding weight. The
  weights for consecutive shift constraint and for consecutive working
  days are respectively 15 and 30.
\item[\textsf{S3}. Consecutive days off (30):] Minimum and maximum number of
consecutive days off should be respected.  
  Their evaluation involves also the border data. Each extra or missing day is 
  multiplied by the corresponding weight.
\item[\textsf{S4}. Preferences (10):] Each assignment to an undesired shift is penalised
  by the corresponding weight.
\item[\textsf{S5}. Complete week-end (30):] Every nurse that has the complete
  weekend value set to $true$, must work both week-end days or none. If
  she/he works only one of the two days \texttt{Sat} and \texttt{Sun}
  this is penalised by the corresponding weight.
\end{description}

\subsubsection{Constraints spanning over the planning horizon}
\label{sec:global-constraints}

The following (soft) constraints are evaluated only at the end of the
planning period:

\begin{description}
\item[\textsf{S6}. Total assignments (20):] For each nurse the total number of
  assignments (working days) must be included within the limits
  (minimum and maximum) enforced by her/his contract.  The difference
  (in either direction), multiplied by its weight, is added to the
  objective function.
\item[\textsf{S7}. Total working week-ends (30):] For each nurse the number of
  working weed-ends must be less than or equal to the maximum. The
  number of worked week-ends in excess is added to the objective
  function multiplied by the weight.  A week-end is considered
  ``working'' if at least one of the two days (\texttt{Sat} and
  \texttt{Sun}) is busy for the nurse.
\end{description}

Obviously, the solver should take constraints \textsf{S6} and
\textsf{S7} into account in each single stage. However, their
violation values have a decreasing degree of uncertainty going from
one week to the following one, and only in the last week they can be
evaluated exactly. It is up to the solver to decide the way to model
them in the cost function in the previous weeks.

\section{Instances}
\label{sec:instances}
One complete solution process requires as input a scenario, an initial
history, and 4 (or 8) week data, and it produces 4 (or 8) week
solutions and one final history. Scenario, week data, history, and
week solutions are written in separate files each one with its own
syntax. 

For ease of processing, all files are provided in XML, JSON, and
text-only formats, and each participant can use the format that he/she
considers as most convenient for his/her implementation.  File formats
are explained in Appendix~\ref{sec:formats}.

Files belonging to the same case are grouped in a \emph{dataset}, which
is composed by the following set of files:
\begin{itemize}
\item 1 scenario file;
\item 3 initial history files;
\item 10 week data files.
\end{itemize}

An \emph{instance} is thus a specific scenario, an initial history, and a
sequence of 4 (or 8) week data files, all belonging to the same dataset.
The same week data file can also be used multiple times in the same instance.

We provide a testbed composed of 14 datasets, one for each combination
of number of weeks and number of nurses, taken from the sets
\{30,40,50,60,80,100,120\} and \{4,8\}, respectively. Datasets are
named using these two number with the prefixes \texttt{n} (for nurses)
and \texttt{w} for weeks. For example, the dataset \texttt{n050w8} is
the (unique) one with 50 nurses and 8 weeks.

In addition, three test datasets, \texttt{n005w4}, \texttt{n012w8}, and
\texttt{n012w4}, are provided for testing and debugging purposes. For
the test datasets, we also provide the solution for three specific instances.

The instances of the competition testbed that will be used for evaluating the
participants will be released on May 15$^{th}$ as the late data.

\section{Tools}
\label{sec:tools}

We provide to the participants a suite of software tools. The
simulator manages the multi-stage solution process. The validator
certificates the quality of a given instance. The benchmark executable
computes the allowed running time for each computer. Finally, the
feasibility checker gives the possibility to the participant to check
that a specific instance of a given dataset has or not at least one
solution that satisfies all hard constraints.

\subsection{Simulator}
\label{sec:simulator}

On the competition website, the \texttt{java} program
\texttt{Simulator.jar} that runs the simulation and delivers the
costs is available. As shown in Figure~\ref{fig:overall-process}, the simulator
receives a scenario file, an initial history file, the solver's
executable file name and a sequence of week data files as its input.
It then applies the solver on each week data file, generating a
history file for each stage based on the solution obtained from each
solver call.  After the last week data file in the sequence is solved,
the validator is called to evaluate the whole planning horizon.
Besides the basic input described above, the user can also specify
random seeds for his/her solver, the directory where the solver is run
in, and the directory where all solution files, generated history
files, log files (the solver's console output after each solver call)
and the validator's results are saved in. In addition to generating a
solution file after solving each stage, the solver might want to save
some other information in a custom file and passing it to
the next solver call, in order to guide the solving of the next stage
better. Such a need is also supported by the simulator.

The simulator is called using the following command line parameters (parameters in the square brackets are optional):
\begin{verbatim} 
java -jar Simulator.jar --sce <Scenario_File> --his <Initial_History_File> 
--weeks <Week_Data_File_1> .. <Week_Data_File_N> --solver <Solver_Executable> 
[--runDir <Running_Directory>] [--outDir <Output_Directory>] [--cus] 
[--rand <Random_Seed_1> .. <Random_Seed_N>]
\end{verbatim}

In detail:

\begin{itemize}
\item File names of the scenario file, initial history files, and week
  data files can have either absolute or relative path. If they are relative,
  they will be taken from the current working directory (not the
  \texttt{Running\_Directory}).

\item The number of week data files \texttt{N} must be equal to the
  number of weeks stated in the scenario file.

\item Before each solver's call, the command line \texttt{cd
    <Running\_Directory>} will be called. 

\item The number of random seeds
  specified after the \texttt{--rand} option can be either one or N.
  If only one random seed is given, it will be used for all solver's
  calls.
\end{itemize}

The simulator works under the assumption that the solver has the following command line:

\begin{verbatim}
Solver_Executable --sce <Scenario_File> --his <Initial_History_File> 
--week <Week_Data_File> --sol <Solution_File_Name> 
[--cusIn <Custom_Input_File>] [--cusOut <Custom_Output_File>] [--rand <Random_Seed>] 
\end{verbatim}

The simulator generates the following files in the directory \texttt{Output\_Directory}:

\begin{itemize}
\item \texttt{history-week0.<extension>, history-week1.<extension>, ..., history-weekN.<extension>}: history files generated by the simulator after each solver call. The \texttt{<extension>} is set based on the relevant input file format.

\item \texttt{custom-week0, custom-week1, ..., custom-weekN}: custom files 
generated by the solver, if the \texttt{--cus} option is used.

\item \texttt{result-week0.txt, result-week1.txt, ..., result-weekN.txt}: log files for all 
solver calls.

\item \texttt{Validator-results.txt}: results of the validator.
\end{itemize}

The simulator asks the solver to generate a single solution file for each stage. These solution files are also located in the directory \texttt{Output\_Directory}, under the names \texttt{sol-week0.<extension>, sol-week1.<extension>, ..., sol-weekN.<extension>}.

Important note: the simulator and the validator must be in the same directory.

As an example, the command line for the simulator

\begin{verbatim}
java -jar Simulator.jar --his H0-n005w4-0.txt --sce Sc-n005w4.txt 
--weeks WD-n005w4-2.txt WD-n005w4-0.txt WD-n005w4-2.txt WD-n005w4-1.txt 
--solver ./program.exe --runDir Devel/ --outDir Simulator_out/ 
--cus --rand 10 11 12 13
\end{verbatim}

produces the following subsequent command lines

\begin{verbatim}
cd Devel/

./program.exe --sce /home/nguyen/data/Sc-n005w4.txt --his /home/nguyen/data/H0-n005w4-0.txt 
--week /home/nguyen/data/WD-n005w4-2.txt --sol /home/nguyen/data/Simulator_out/sol-week0.txt 
--cusOut /home/nguyen/data/Simulator_out/custom-week0 --rand 10

./program.exe --sce /home/nguyen/data/Sc-n005w4.txt 
--his /home/nguyen/data/Simulator_out/history-week0.txt 
--week /home/nguyen/data/WD-n005w4-0.txt --sol /home/nguyen/data/Simulator_out/sol-week1.txt 
--cusIn /home/nguyen/data/Simulator_out/custom-week0 
--cusOut /home/nguyen/data/Simulator_out/custom-week1 --rand 11

./program.exe --sce /home/nguyen/data/Sc-n005w4.txt 
--his /home/nguyen/data/Simulator_out/history-week1.txt 
--week /home/nguyen/data/WD-n005w4-2.txt --sol /home/nguyen/data/Simulator_out/sol-week2.txt 
--cusIn /home/nguyen/data/Simulator_out/custom-week1 
--cusOut /home/nguyen/data/Simulator_out/custom-week2 --rand 12

./program.exe --sce /home/nguyen/data/Sc-n005w4.txt 
--his /home/nguyen/data/Simulator_out/history-week2.txt 
--week /home/nguyen/data/WD-n005w4-1.txt --sol /home/nguyen/data/Simulator_out/sol-week3.txt 
--cusIn /home/nguyen/data/Simulator_out/custom-week2 
--cusOut /home/nguyen/data/Simulator_out/custom-week3 --rand 13
\end{verbatim}

Given that the current working directory is \texttt{/home/nguyen/data/}

\subsection{Validator}

The validator is a
\texttt{java} program that checks for the validity of a solution of an
instance, and calculate the corresponding objective function value
according to the evaluation method (see also 
Appendix~\ref{sec:evaluation}) and the constraints' weights described
in Section~\ref{subsec:full_solution_process}.  The validator is
automatically called by the simulator at the end of the solving
procedure. It can also be used as a stand-alone program with the
following syntax:

\begin{verbatim}
java -jar Validator.jar --sce <Scenario_File> --his <Initial_History_File> 
--weeks <Week_Data_File_1> .. <Week_Data_File_N> 
--sols <Solution_File_1> .. <Solution_File_N> [--verbose]
\end{verbatim}

If the \texttt{--verbose} option is used, details of each soft
constraint's violation of each nurse are shown. As an example, the command line

\begin{verbatim} 
java -jar validator.jar --sce n005w4/Sc-n005w4.txt --his n005w4/H0-n005w4-0.txt 
--weeks n005w4/WD-n005w4-1.txt n005w4/WD-n005w4-2.txt n005w4/WD-n005w4-3.txt 
n005w4/WD-n005w4-3.txt --sols Solutions/Sol-n005w4-1-0.txt Solutions/Sol-n005w4-2-1.txt 
Solutions/Sol-n005w4-3-2.txt Solutions/Sol-n005w4-3-3.txt
\end{verbatim}

produces an output like:

\begin{small}
\begin{verbatim} 
        |M|T|W|T|F|S|S| |M|T|W|T|F|S|S| |M|T|W|T|F|S|S| |M|T|W|T|F|S|S|
-------------------------------------------------------------------------
Patrick |N|-|E|E|E|L|L| |-|-|E|E|L|L|L| |-|N|N|N|N|N|N| |-|L|L|L|L|N|N|
Andrea  |L|L|-|-|L|L|L| |N|N|N|N|N|-|L| |L|L|L|-|-|N|N| |N|N|N|-|-|E|E|
Stefaan |N|N|N|N|-|-|-| |E|E|L|L|-|-|E| |N|N|-|-|E|E|E| |N|N|-|-|-|L|L|
Sara    |-|-|-|N|N|N|N| |N|-|-|-|E|E|E| |E|L|L|L|-|-|-| |E|E|E|E|E|-|-|
Nguyen  |E|E|L|L|-|E|E| |L|L|-|L|N|N|N| |-|E|E|E|L|L|L| |-|L|L|N|N|N|N|

Hard constraint violations
--------------------------
Minimal coverage constraints: 0
Required skill constraints: 0
Illegal shift type succession constraints: 0
Single assignment per day: 0

Cost per constraint type
------------------------
Total assignment constraints: 320
Consecutive constraints: 465
Non working days constraints: 330
Preferences: 70
Max working weekend: 210
Complete weekends: 60
Optimal coverage constraints: 240

------------------------
Total cost: 1695
\end{verbatim}
\end{small}

\subsection{Benchmark}
\label{sec:benchmarking}

The benchmark program is designed to test how fast your machine is at
doing the sort of things that are involved in rostering.  For each
problem size, which is defined as the number of nurses
in the scenario file, the program tells you how long you can run
your algorithm for each stage. It is not possible to provide perfectly
equitable benchmarks across many platforms and algorithms, and we know
that the benchmark may be kinder to some people than others. It is
pointed out that all the finalists will be run on a standard machine
therefore creating a `level playing field'.

The benchmark is only suitable for individual, single processor
machines. It is not suitable, for example, for specialist parallel
machines or clusters. In general, for multi-core machines, one single
core is allowed to be used for the competition.

The benchmark is provided as an executable for various architectures.
If your architecture is not among the ones provided please contact us to obtain the program.

The program should be run when the machine is not being used for
anything else.  The program will report how long it took, and hence
the length of time you can run your rostering algorithm per stage, for
each number of nurses.

On a relatively modern PC, the benchmark program will grant the
participant approximately $10 + 30 * (N-20)$ seconds for each stage,
in which $N$ is the number nurses.

\subsection{Feasibility Checker}

It could be possible that some instance created from a given dataset
is infeasible. In order to prevent participants from wasting time for
searching for feasible solutions when they do not exists, a feasibility
checker is provided. This tool is a web service and can be found on the
competition website from November 5, 2014.

\section{Competition rules}
\label{sec:rules}

\subsection{General rules}

This competition seeks to encourage research into automated nurse
rostering methods for solving a multi-stage nurse rostering problem,
and to offer prizes to the most successful methods. It is the spirit
of these rules that is important, not the letter. With any set of
rules for any competition it is possible to work within the letter of
the rules but outside the spirit.

\begin{description}
\item[\textbf{Rule 1:}] The organisers reserve the right to disqualify
  any participant from the competition at any time if the participant
  is determined by the organisers to have worked outside the spirit of
  the competition rules. The organisers' decision is final in any
  matter. 
\item[\textbf{Rule 2:}] The organisers reserve the right to change the
  rules at any time, if they believe it is necessary for the sake of
  preserving the correct operation of the competition. Any change of
  rules will be accompanied by a general email to all participants.

\item[\textbf{Rule 3:}] The competition has a deadline when all
  submissions must be uploaded. The deadline is strict and no
  extensions will be given under any circumstances.

\item[\textbf{Rule 4:}] Participants can use any programming
  language. The use of third-party software is allowed under the
  following restrictions:
\begin{itemize}
    \item it is free software;
    \item it's behaviour is (reasonably) documented;
    \item it runs under a commonly-used operating system (Unix/Linux, Windows, or Mac OS X).
\end{itemize}

\item[\textbf{Rule 5:}] Participants have to benchmark their machine
  with the program provided in order to know how much time they have
  available to run their program for each stage on their machines. The
  solver should run on a single core of the machine.

\item[\textbf{Rule 6:}] The solver should take as input the
  files in one of the formats described, and produce as output a list of
  solution files for all stages (in the same format). It should do
  so within the allowed CPU time. 

\item[\textbf{Rule 7:}] The solver used should be the same executable
  for all weeks, but obviously internally it could exploit the
  information regarding the week it is solving, and adapt its behavior
  depending on it.

\item[\textbf{Rule 8:}] The solver can be either deterministic or
  stochastic. In both cases, participants must be prepared to show
  that the results are repeatable in the given computer time. In
  particular, the participants that use a stochastic algorithm should
  code their program in such a way that the exact run that produced
  each solution submitted can be repeated (by recording the random
  seed). 

\item[\textbf{Rule 9:}] Along with the solution for each instance,
  the participants should also submit a concise and clear description
  of their algorithm, so that in principle others can implement it.

\item[\textbf{Rule 10:}] A set of 5 finalists will be chosen after the
  competition deadline. Ordering of participants will be based on the
  scores obtained on the provided instances. The actual list will be based
  on the ranks of solvers on each single instance. The mean average of
  the ranks will produce the final place list. More details on how the
  orderings will be established can be found in Section
  \ref{subsec:Adjudication}.

\item[\textbf{Rule 11:}] The finalists will be asked to provide the
  executable that will be run and tested by the organisers. The
  finalists' solvers will be rerun by the organisers on new instances
  (including new datasets). It is the responsibility of the
  participant to ensure all information is provided to enable the
  organisers to recreate the solution.  If appropriate information is
  not received or indeed the submitted solutions cannot be recreated,
  another finalist will be chosen from the original participants.

\item[\textbf{Rule 12:}] Finalists' eventual place listings will be
  based on the ranks on each single instance for a set of trials on
  all instances. As with Rule 10, an explanation of the procedures to
  be used can be found in Section \ref{subsec:Adjudication}.

\item[\textbf{Rule 13:}] In some circumstances, finalists may be
  required to show source code to the organisers. This is simply to
  check that they have stuck to the rules and will be treated in the
  strictest confidence.

\item[\textbf{Rule 14:}] Organisers of the competition cannot participate to it. 
\end{description}

\subsection{Dates}

The competition starts on October 17, 2014. On this date, we
release the datasets, this specification paper, the simulator, the validator and the benchmark program. The web service for the feasibility checker is available on November 5, 2014. On May 15, 2015, the late data,
i.e., a list of specific instances created from the competition
testbed, will be released. The deadline for submission of participants'
best results and their solvers is June 1, 2015. Notifications of the
finalists will be sent out on July 1, 2015. The winners will be
announced at the MISTA 2015 Conference in Prague (August 25-27, 2015).

\subsection{Adjudication procedure}
\label{subsec:Adjudication}

We follow the same adjudication procedure of INRC-I \cite{HDSS12},
from which in turn has been imported from the Second International
Timetabling Competition (ITC-2007) \cite{MSPM10}. It is repeated here
for the sake on selfcontainedness.

Let $m$ be the total number of instances and $k$ be the number of
participants. Let $X_{ij}$ be the value of the objective function $s$
supplied (and verified) by participant $i$ for instance $j$. In case
participant $i$ provides an infeasible solution for instance $j$ or
he/she does not provide it at all, $X_{ij}$ is assigned a conventional
value $M$ larger than all the results supplied by the other
participants for that instance.

The matrix $X$ of results is transformed into a matrix of ranks $R$
assigning to each $R_{ij}$ a value from $1$ to $k$. That is, for
instance $j$ the supplied $X_{1j}$, $X_{2j}$, \ldots ,$X_{kj}$ are
compared with each other and the rank $1$ is assigned to the smallest
observed value, the rank $2$ to the second smallest, and so on to the
rank $k$, which is assigned to the largest value for instance $ij$.
We use average ranks in case of ties.

 
Consider the example with $m=6$ instances and $k=7$ participants in
Table \ref{tbl:sols}. The ranks are shown in Table \ref{tbl:ranks}.

\begin{table}[H]
\centering
	\begin{tabular}{|l|r|r|r|r|r|r|}
	\hline
	Instance & 1 & 2 & 3 & 4 & 5 & 6 \\
	\hline
	Solver 1 & 34 & 35 & 42 & 32 & 10 & 12\\
	\hline
	Solver 2 & 32 & 24 & 44 & 33 & 13 & 15\\
	\hline
	Solver 3 & 33 & 36 & 30 & 12 & 10 & 17\\
	\hline
	Solver 4 & 36 & 32 & 46 & 32 & 12 & 13\\
	\hline
	Solver 5 & 37 & 30 & 43 & 29 & 9 & 4\\
	\hline
	Solver 6 & 68 & 29 & 41 & 55 & 10 & 5\\
	\hline
	Solver 7 & 36 & 30 & 43 & 58 & 10 & 4\\
	\hline
	\end{tabular}
	\caption{An example of scores of the submitted solutions.}
	\label{tbl:sols}
\end{table}

\begin{table}[h]
\centering
	\begin{tabular}{|l|r|r|r|r|r|r|}
	\hline
	Instance & 1 & 2 & 3 & 4 & 5 & 6 \\
	\hline
	Solver 1 & 3 & 6 & 3 & 3.5 & 3.5 & 4\\
	\hline
	Solver 2 & 1 & 1 & 6 & 5 & 7 & 6\\
	\hline
	Solver 3 & 2 & 7 & 1 & 1 & 3.5 & 7\\
	\hline
	Solver 4 & 4.5 & 5 & 7 & 3.5 & 6 & 5\\
	\hline
	Solver 5 & 6 & 3.5 & 4.5 & 2 & 1 & 1.5\\
	\hline
	Solver 6 & 7 & 2 & 2 & 6 & 3.5 & 3\\
	\hline
	Solver 7 & 4.5 & 3.5 & 4.5 & 7 & 3.5 & 1.5\\
	\hline
	\end{tabular}
	\caption{Corresponding solution ranks for the example.}
	\label{tbl:ranks}
\end{table}
 
We define for each solver the mean of the ranks. The finalists of the
competition will be the 5 solvers with the lowest mean ranks. In case
of a tie for entering the last positions, all the last equal-mean
solvers are included in the final (in this case the finalists will be
more than 5). In the example, the mean ranks are shown in Table
\ref{tbl:meanranks}. In this case the finalists would be solvers 1, 3, 5, 6 and 7.

\begin{table}[h]
\centering
	\begin{tabular}{|l|r|}
	\hline
	Solver 1 & 3.83 \\
	\hline
	Solver 2 & 4.33 \\
	\hline
	Solver 3 & 3.58 \\
	\hline
	Solver 4 & 5.17 \\
	\hline
	Solver 5 & 3.08 \\
	\hline
	Solver 6 & 3.92 \\
	\hline
	Solver 7 & 4.08 \\
	\hline
	\end{tabular}
	\caption{Mean ranks.}
	\label{tbl:meanranks}
\end{table}

The organisers will check the runs of the candidate finalist with the
submitted seed to make sure that the submitted runs are repeatable. If
they are not, then another entrant will be chosen for the final.
 
For the final, the same evaluation process is repeated for the
finalists with the following differences:

\begin{enumerate} 
\item New instances, including hidden datasets, will be used.
\item The solvers will be run by the organisers, thus the finalist
  should give support to the organisers in the process of compiling
  and running the solvers.
\item For each instance, the organisers will run 10 independent trails
  with seeds chosen at random. For each trial, we will compute the
  ranks and average them on all trials on all instances.
\end{enumerate}
 
The winner is the one with the lowest mean rank. In case of a tie, 1
trial is added for all instances until a single winner is found.

\subsection{Prizes}
The top three will divide \EUR{1729} among them (first prize
\EUR{819}, second \EUR{637}, third \EUR{273}), and will be offered
free registration to PATAT 2016, that will include a special track on
the competition.

\appendix

\section{File formats}
\label{sec:formats}

In this appendix we describe the format for the input files (scenario,
week data, and history) and output file (solution). Only the text-only
format is explained in details, given that XML and JSON files are organised
with the same structure, but in a more self-explanatory way.

\subsection*{Scenario}

The first line of the scenario file contains the name of the dataset
in the format \texttt{nXXXwY}, where \texttt{XXX} is the number of
nurses and \texttt{Y} the number of weeks of the planning horizon.
This is the identifier of the scenario that is subsequently used in
the relating history, week data, and solution files.

\begin{small}
\begin{verbatim}
SCENARIO = n005w4 
\end{verbatim}
\end{small}

Then it is reported the length of the planning horizon, expressed in
number of weeks, and the number and names of skills for nurses.

\begin{small}
\begin{verbatim}
WEEKS = 4

SKILLS = 2
HeadNurse
Nurse
\end{verbatim}
\end{small}

The shift types section indicates the number of shift types available,
and for each one, the identifier (its name), and the minimum and the
maximum number of consecutive assignments allowed. For each shift
type, it is also detailed the forbidden shift types sequences as
\texttt{$\langle$preceding\_shift\_type$\rangle$}
\texttt{$\langle$number\_forbidden\_successions$\rangle$}
\texttt{$\langle$succeeding\_shift\_type\_list$\rangle$}.  In the
following example, the successions \texttt{Late $\rightarrow$ Early},
\texttt{Night $\rightarrow$ Early} and \texttt{Night $\rightarrow$
  Late} are forbidden.

\begin{small}
\begin{verbatim}
SHIFT_TYPES = 3
Early (2,5)
Late (2,3)
Night (4,5)

FORBIDDEN_SHIFT_TYPES_SUCCESSIONS
Early 0
Late 1 Early
Night 2 Early Late
\end{verbatim}
\end{small}

In the contract section it is listed the name of the contract type,
and the lower and upper limits on working and rest days. In detail, it
establishes the minimum and the maximum number of total assignments in
the planning horizon, the minimum and the maximum number of
consecutive working days, the minimum and the maximum number of
consecutive days off, the maximum number of working weekends, and the
presence (\texttt{1}) or absence (\texttt{0}) of the complete weekend
constraint.

\begin{small}
\begin{verbatim}
CONTRACTS = 2
FullTime (15,22) (3,5) (2,3) 2 1
PartTime (7,11) (3,5) (3,5) 2 1
\end{verbatim}
\end{small}

Finally, the nurse section reports the total number of nurses available, and for each nurse his/her identifier (the name), the contract type, the number of skills owned and their names.

\begin{small}
\begin{verbatim}
NURSES = 5
Patrick FullTime 2 HeadNurse Nurse 
Andrea FullTime 2 HeadNurse Nurse
Stefaan PartTime 2 HeadNurse Nurse
Sara PartTime 1 Nurse
Nguyen FullTime 1 Nurse
\end{verbatim}
\end{small}

\subsection*{Week data}

In the week data file, first of all there is the identifier of the
corresponding scenario; then all the data about coverage requirements
and nurse preferences is listed.

\begin{small}
\begin{verbatim}
WEEK_DATA
n005w4
\end{verbatim}
\end{small}

A coverage requirement is specified by the shift type, the skill, and for each day of the week (from Monday to Sunday), the minimum coverage and the optimal coverage.

\begin{small}
\begin{verbatim}
REQUIREMENTS
Early HeadNurse (1,1) (0,0) (0,0) (0,0) (0,0) (1,1) (0,0)
Early Nurse (1,2) (1,1) (1,1) (0,1) (1,1) (1,1) (0,1)
Late HeadNurse (1,1) (0,1) (1,1) (0,0) (0,0) (0,0) (0,0)
Late Nurse (1,1) (1,1) (0,1) (0,1) (1,1) (1,1) (1,1)
Night HeadNurse (0,0) (1,1) (0,0) (0,0) (1,1) (1,1) (0,0)
Night Nurse (0,1) (1,1) (1,1) (1,1) (1,1) (0,1) (1,1)
\end{verbatim}
\end{small}

Finally, the number of shift off requests is reported with the following grammar: \texttt{$\langle$nurse$\rangle$} \texttt{$\langle$shift type$\rangle$} \texttt{$\langle$day$\rangle$}. The special shift type \texttt{Any} means that the nurse would like to have a day off.

\begin{small}
\begin{verbatim}
SHIFT_OFF_REQUESTS = 3
Sara Any Thu 
Sara Night Sat 
Stefaan Late Sat 
\end{verbatim}
\end{small}

\subsection*{History}

The first line of the history file describes the week to which the
history refers to (i.e. $0$ for the initial history file, $1$ after
the first week, \dots) and the relating scenario file.

\begin{small}
\begin{verbatim}
HISTORY
0 n005w4
\end{verbatim}
\end{small}

In addition, the file contains the nurse history, in terms of total
number of assignments, total number of worked weekends, last assigned
shift type, number of consecutive assignments of the last shift type,
number of consecutive worked days and number of consecutive days off.

\begin{small}
\begin{verbatim}
NURSE_HISTORY
Patrick 0 0 Night 1 4 0
Andrea 0 0 Early 3 3 0
Stefaan 0 0 None 0 0 3
Sara 0 0 Late 1 4 0
Nguyen 0 0 None 0 0 1
\end{verbatim}
\end{small}

\subsection*{Solution}
The solution file gives the assignment of nurses to shifts and skills.
The file starts with the reference to the solved week and to the scenario.

\begin{small}
\begin{verbatim}
SOLUTION
3 n005w4
\end{verbatim}
\end{small}

Then each single assignment is shown (in any order), reporting the
name of the nurse, the day, the shift type and the skill considered.
Days off are neglected.

\begin{small}
\begin{verbatim}
ASSIGNMENTS = 26
Patrick Mon Late HeadNurse
Patrick Tue Night HeadNurse
Patrick Fri Early Nurse
Patrick Sat Early Nurse
Patrick Sun Late Nurse
Andrea Mon Early HeadNurse
Andrea Tue Late Nurse
..
Nguyen Fri Late Nurse
Nguyen Sat Late Nurse
Nguyen Sun Night Nurse
\end{verbatim}
\end{small}

\section{Constraint evaluation}
\label{sec:evaluation}

In this appendix, we explain in more detail some of the constraints
presented in Section~\ref{sec:constraints}. Specifically, we believe that the
constraints that involve border data (history file) need a deeper
explanation.  Conversely, constraints \textsf{H1}, \textsf{H2},
\textsf{S1}, \textsf{S4}, \textsf{S5}, \textsf{S6}, and \textsf{S7} do
not rely on border data for their evaluation, and, in our opinion,
their evaluation is straightforward and not subject to ambiguous
interpretation.

For the remaining constraints, we exhaustively describe their evaluation
at both \emph{borders} of a stage. 

In the descriptions that follow, for simplicity, we omit the weight of
the (soft) constraints, and we focus on the amount of the violation.
For simplicity, we assume that the input data is the one included in
the following fragment of a scenario file, setting all limits to the
same value 3.

\begin{verbatim}
SHIFT_TYPES = 2
Early (3,3)
Late (3,3)

FORBIDDEN_SHIFT_TYPES_SUCCESSIONS
Early 0
Late 1 Early

CONTRACTS = 1
FullTime (...,...) (3,3) (3,3) ... 
\end{verbatim}


Section~\ref{sec:consworkday} explains the number of consecutive
working days constraints (\textsf{S2}). In
section~\ref{sec:consdayoff} the evaluation of the number of
consecutive days off constraints is explained (\textsf{S3}).
Section~\ref{sec:shiftsuccs} elaborates  on the forbidden shift type
successions constraint (\textsf{H3}).

\subsection{Number of consecutive assignments (\textsf{S2})}
\label{sec:consworkday}
 
\subsubsection*{Maximum number of consecutive working days}

The evaluation of the constraint at the start of a stage depends on
the value $c$ of the number of consecutive working days at the
beginning of a stage (from history).  In
Table~\ref{tbl:eval:maxcwdbegin}, the constraint is evaluated for
$c=5$, showing for clarity also the week before the planning period.

The symbol \W{} is used to mean any working shift, and the symbol \D{}
means a day off. An empty cell is used for assignments irrelevant for
the example under consideration.

As it can be seen, the evaluator only counts the `extra' amount of
violation, as part of the violation has already been taken into
account during the evaluation of the previous stage (see
table~\ref{tbl:eval:maxcwdend}).


\begin{table}[H]
  \centering
  \begin{tabular}{|c|c|c|c|c|c|c||c|c|c|c|c|c|c|c|}
  \cline{1-14}
  \multicolumn{7}{|c||}{Previous period} & \multicolumn{7}{|c|}{Current period} & \multicolumn{1}{c}{} \\ \hline
   Mo &  Tu & We & Th & Fr & Sa & Su & Mo & Tu & We & Th & Fr & Sa & Su & Violations \\ \hline
      & \D & \W  & \W  & \W  & \W  & \W  & \D   &    &    &    &    &    &    & 0 \\ \hline
      & \D & \W  & \W  & \W  & \W  & \W  & \W  & \D   &    &    &    &    &    & 1 \\ \hline
      & \D & \W  & \W  & \W  & \W  & \W  & \W  & \W  & \D   &    &    &    &    & 2 \\ \hline  
  \end{tabular}
  \caption{Evaluation of the maximum number of consecutive working days constraint  at the beginning of a stage.}
  \label{tbl:eval:maxcwdbegin}
\end{table}


From this point on, the previous planning period will
be represented by a single column denoting the value of the relevant
counter from history. 
Table~\ref{tbl:eval:maxcwdbeginall} shows the evaluation of the
constraint for different values of $c$.

\begin{table}[H]
  \centering
  \begin{tabular}{|c|c|c|c|c|c|c|c|c|}
  \hline
                    History & Mo & Tu & We & Th & Fr & Sa & Su & Violations \\ \hline
   \multirow{5}{*}{$c \ge 3$} & \D   &    &    &    &    &    &    & 0 \\ \cline{2-9}
                            & \W  & \D   &    &    &    &    &    & 1 \\ \cline{2-9}
                            & \W  & \W  & \D   &    &    &    &    & 2 \\ \cline{2-9}
                            & \W  & \W  & \W  & \D   &    &    &    & 3 \\ \cline{2-9}
                            & \W  & \W  & \W  & \W  & \D   &    &    & 4 \\ \hline \hline
   \multirow{5}{*}{$c = 2$} & \D  &    &    &    &    &    &    & 0 \\ \cline{2-9}
                            & \W  & \D  &    &    &    &    &    & 0 \\ \cline{2-9}
                            & \W  & \W  & \D  &    &    &    &    & 1 \\ \cline{2-9}
                            & \W  & \W  & \W  & \D    &    &    &    & 2 \\ \cline{2-9}
                            & \W  & \W  & \W  & \W  &\D     &    &    & 3 \\ \hline \hline
   \multirow{5}{*}{$c = 1$} & \D   &    &    &    &    &    &    & 0 \\ \cline{2-9}
                            & \W  & \D   &    &    &    &    &    & 0 \\ \cline{2-9}
                            & \W  & \W  &  \D  &    &    &    &    & 0 \\ \cline{2-9}
                            & \W  & \W  & \W  &  \D  &    &    &    & 1 \\ \cline{2-9}
                            & \W  & \W  & \W  & \W  & \D   &    &    & 2 \\ \hline \hline
   \multirow{5}{*}{$c = 0$} & \D   &    &    &    &    &    &    & 0 \\ \cline{2-9}
                            & \W  &  \D  &    &    &    &    &    & 0 \\ \cline{2-9}
                            & \W  & \W  &  \D  &    &    &    &    & 0 \\ \cline{2-9}
                            & \W  & \W  & \W  &  \D  &    &    &    & 0 \\ \cline{2-9}
                            & \W  & \W  & \W  & \W  &  \D  &    &    & 1 \\ \hline 
  \end{tabular}
  \caption{Evaluation of the maximum number of consecutive working days constraint at the beginning of a stage for different values of $c$.}
  \label{tbl:eval:maxcwdbeginall}
\end{table}

Table~\ref{tbl:eval:maxcwdend} shows the evaluation of the constraint
for the maximum consecutive working days at the end of a stage.

\begin{table}[H]
  \centering
  \begin{tabular}{|c|c|c|c|c|c|c|c|}
  \hline
   Mo & Tu & We & Th & Fr & Sa & Su & Violations \\ \hline
   \D   & \W  & \W  & \W  & \W  & \W  & \W  & 3 \\ \hline
      & \D   & \W  & \W  & \W  & \W  & \W  & 2 \\ \hline
      &    & \D   & \W  & \W  & \W  & \W  & 1 \\ \hline
      &    &    &  \D  & \W  & \W  & \W  & 0 \\ \hline
  \end{tabular}
  \caption{Evaluation of the maximum number of consecutive working days constraint  at the end of a stage.}
  \label{tbl:eval:maxcwdend}
\end{table}

\subsubsection*{Minimum number of consecutive working days}

Table~\ref{tbl:eval:mincwd} evaluates the constraint for the minimum
consecutive working days. If $c\ge 3$, then no violation of this
constraint can occur at the beginning of a stage.  As it is uncertain
what the assignments at the beginning of the next stage are, the
minimum number of consecutive working days constraint is not taken
into account at the end of a stage.

\begin{table}[H]
  \centering
  \begin{tabular}{|c|c|c|c|c|c|c|c|c|}
  \hline
                    History & Mo & Tu & We & Th & Fr & Sa & Su & Violations \\ \hline
   \multirow{2}{*}{c = 2}   & \D  &    &    &    &    &    &    & 1 \\ \cline{2-9}
                            & \W  & \D &    &    &    &    &    & 0 \\ \hline\hline
   \multirow{3}{*}{c = 1}   & \D  &    &    &    &    &    &    & 2 \\ \cline{2-9}
                            & \W  & \D &    &    &    &    &    & 1 \\ \cline{2-9}
                            & \W  & \W & \D &    &    &    &    & 0 \\ \hline\hline
   \multirow{4}{*}{c = 0}   & \D  &    &    &    &    &    &    & 0 \\ \cline{2-9}
                            & \W  & \D &    &    &    &    &    & 2 \\ \cline{2-9}
                            & \W  & \W & \D &    &    &    &    & 1 \\ \cline{2-9}
                            & \W  & \W & \W & \D &    &    &    & 0 \\ \hline
  \end{tabular}
  \caption{Evaluation of the minimum number of consecutive working days constraint at the beginning of a stage.}
  \label{tbl:eval:mincwd}
\end{table}

Note that both the maximum and minimum constraints are evaluated per
series. In Table~\ref{tbl:eval:multiseries}, there are two series of
length 1 and 2, respectively, that produce two and one violations,
respectively (minimum is 3).

\begin{table}[H]
  \centering
  \begin{tabular}{|c|c|c|c|c|c|c|c|}
  \hline
       Mo & Tu & We & Th & Fr & Sa & Su & Violations \\ \hline 
       \D & \W  &  \D  & \D & \W  & \W  & \D   & 3  \\ \hline
  \end{tabular}
  \caption{Evaluation of the minimum number of consecutive working days constraint for two series.}
  \label{tbl:eval:multiseries}
\end{table}

\subsubsection*{Minimum and maximum number of consecutive assignments to the same shift}

The examples presented in the previous paragraphs work similarly for
the consecutive assignments to the same shift, by replacing the symbol \W{}
with the specific shift (E or L, in this example) and the symbol \D{}
with a day off or any shift different from the given one.

\subsection{Number of consecutive days off (\textsf{S3})}
\label{sec:consdayoff}

\subsubsection{Maximum number of consecutive days off}

The evaluation of the constraint at the start of a stage depends
on the value of the number of consecutive days off from the history, that we
call $c$ again. Table~\ref{tbl:eval:maxcdobegin} shows the evaluation of the maximum
number of consecutive days off constraint at the beginning of the
stage.

\begin{table}[H]
  \centering
  \begin{tabular}{|c|c|c|c|c|c|c|c|c|}
  \hline
                    History & Mo & Tu & We & Th & Fr & Sa & Su & Violations \\ \hline
   \multirow{5}{*}{$c \ge 3$} & \W &    &    &    &    &    &    & 0 \\ \cline{2-9}
                            & \D & \W &    &    &    &    &    & 1 \\ \cline{2-9}
                            & \D & \D & \W &    &    &    &    & 2 \\ \cline{2-9}
                            & \D & \D & \D & \W &    &    &    & 3 \\ \cline{2-9}
                            & \D & \D & \D & \D & \W &    &    & 4 \\ \hline\hline 
   \multirow{5}{*}{$c = 2$} & \W &    &    &     &    &    &    & 0 \\ \cline{2-9}
                            & \D & \W &    &     &    &    &    & 0 \\ \cline{2-9}
                            & \D & \D & \W &     &    &    &    & 1 \\ \cline{2-9}
                            & \D & \D & \D & \W  &    &    &    & 2 \\ \cline{2-9}
                            & \D & \D & \D & \D  & \W &    &    & 3 \\ \hline \hline
   \multirow{5}{*}{$c = 1$} & \W &    &    &     &    &    &    & 0 \\ \cline{2-9}
                            & \D & \W &    &     &    &    &    & 0 \\ \cline{2-9}
                            & \D & \D & \W &     &    &    &    & 0 \\ \cline{2-9}
                            & \D & \D & \D & \W  &    &    &    & 1 \\ \cline{2-9}
                            & \D & \D & \D & \D  & \W  &    &    & 2 \\ \hline \hline
   \multirow{5}{*}{$c = 0$} & \W  &     &     &     &    &    &    & 0 \\ \cline{2-9}
                            &\D  & \W  &     &     &    &    &    & 0 \\ \cline{2-9}
                            & \D & \D & \W  &     &    &    &    & 0 \\ \cline{2-9}
                            & \D & \D & \D & \W  &    &    &    & 0 \\ \cline{2-9}
                            & \D & \D & \D & \D & \W  &    &    & 1 \\ \hline 
  \end{tabular}
  \caption{Evaluation of the maximum number of consecutive days off constraint at the beginning of a stage for different values of $c$.}
  \label{tbl:eval:maxcdobegin}
\end{table}

In Table~\ref{tbl:eval:maxcdoend}, we show the violations of the
maximum number of consecutive days off constraint at the end of a
stage; the history is not involve in this case.

\begin{table}[H]
  \centering
  \begin{tabular}{|c|c|c|c|c|c|c|c|}
  \hline
   Mo & Tu & We & Th & Fr & Sa & Su & Violations \\ \hline
      &    &   & \W  & \D & \D & \D & 0 \\ \hline
      &    & \W  & \D & \D & \D & \D & 1 \\ \hline
      & \W & \D & \D & \D & \D & \D & 2 \\ \hline
   \W & \D & \D & \D & \D & \D & \D & 3 \\ \hline
  \end{tabular}
  \caption{Evaluation of the maximum number of consecutive days off constraint at the end of a stage.}
  \label{tbl:eval:maxcdoend}
\end{table}

\subsubsection{Minimum number of consecutive days off}

Table~\ref{tbl:eval:mincdobegin} shows the evaluation of the minimum
number of consecutive days off constraint at the
beginning of the stage. 

\begin{table}[H]
  \centering
  \begin{tabular}{|c|c|c|c|c|c|c|c|c|}
  \hline
                    History & Mo & Tu & We & Th & Fr & Sa & Su & Violations \\ \hline
   \multirow{5}{*}{$\ge 3$} & \W  &    &    &    &    &    &    & 0 \\ \cline{2-9}
                            & \D & \W  &    &    &    &    &    & 0 \\ \cline{2-9}
                            & \D & \D & \W  &    &    &    &    & 0 \\ \cline{2-9}
                            & \D & \D & \D & \W  &    &    &    & 0 \\ \cline{2-9}
                            & \D & \D & \D & \D & \W  &    &    & 0 \\ \hline \hline 
   \multirow{5}{*}{2}       & \W  &    &    &    &    &    &    & 1 \\ \cline{2-9}
                            & \D & \W  &    &    &    &    &    & 0 \\ \cline{2-9}
                            & \D & \D & \W  &    &    &    &    & 0 \\ \cline{2-9}
                            & \D & \D & \D & \W  &    &    &    & 0 \\ \cline{2-9}
                            & \D & \D & \D & \D & \W  &    &    & 0 \\ \hline \hline 
   \multirow{5}{*}{1}       & \W  &    &    &    &    &    &    & 2 \\ \cline{2-9}
                            & \D & \W  &    &    &    &    &    & 1 \\ \cline{2-9}
                            & \D & \D & \W  &    &    &    &    & 0 \\ \cline{2-9}
                            & \D & \D & \D & \W  &    &    &    & 0 \\ \cline{2-9}
                            & \D & \D & \D & \D & \W  &    &    & 0 \\ \hline \hline
   \multirow{5}{*}{0}       & \W  &    &    &    &    &    &    & 0 \\ \cline{2-9}
                            & \D  & \W  &    &    &    &    &    & 2 \\ \cline{2-9}
                            & \D  & \D & \W  &    &    &    &    & 1 \\ \cline{2-9}
                            & \D  & \D & \D  & \W  &    &    &    & 0 \\ \cline{2-9}
                            & \D  & \D & \D  & \D & \W  &    &    & 0 \\ \hline 
  \end{tabular}
  \caption{Evaluation of the minimum number of consecutive days off constraint at the beginning of a stage for different values of $c$.}
  \label{tbl:eval:mincdobegin}
\end{table}

\subsection{Forbidden shift type successions}
\label{sec:shiftsuccs}

In Table~\ref{tbl:eval:fbstsucc}, the evaluation of the constraint for
the forbidden succession at the beginning of the stage is shown. 
As can be seen, the constraint is violated only when the last assigned
shift type from history is equal to \texttt{Late} and the Monday
shift is \texttt{Early}.

\begin{table}[H]
  \centering
  \begin{tabular}{|c|c|c|c|c|c|c|c|c|}
  \hline
                    History & Mo & Tu & We & Th & Fr & Sa & Su & Violation \\ \hline
   \multirow{3}{*}{L} & E  &    &    &    &    &    &    & true \\ \cline{2-9}
                      & L  &    &    &    &    &    &    & false \\ \cline{2-9}
                      & \D &  &    &    &    &    &      & false \\ \hline \hline 
   \multirow{3}{*}{E} & E  &    &    &    &    &    &    & false \\ \cline{2-9}
                      & L  &    &    &    &    &    &    & false \\ \cline{2-9}
                      & \D & &    &    &    &    &       & false \\ \hline
  \end{tabular}
  \caption{Evaluation of the forbidden shift type succession L-E.}
  \label{tbl:eval:fbstsucc}
\end{table}
\bibliography{nurse}
\bibliographystyle{plain}

\end{document}